\documentclass[letterpaper]{article} 
\usepackage{aaai22}  
\usepackage{times}  
\usepackage{helvet}  
\usepackage{courier}  
\usepackage[hyphens]{url}  
\usepackage{graphicx} 
\usepackage{natbib}  
\usepackage{caption} 
\usepackage{amsfonts}
\DeclareCaptionStyle{ruled}{labelfont=normalfont,labelsep=colon,strut=off} 
\usepackage{algorithm}
\usepackage{algorithmic}
\usepackage{amsmath}

\usepackage{array}
\usepackage{multirow}
\usepackage{booktabs}
\usepackage{url}
\usepackage[toc,page]{appendix}

\usepackage{newfloat}
\usepackage{listings}
\floatstyle{ruled}
\newfloat{listing}{tb}{lst}{}
\floatname{listing}{Listing}
\title{TransMEF: A Transformer-Based Multi-Exposure Image Fusion Framework using Self-Supervised Multi-Task Learning}

\author{Linhao Qu\thanks{Equal Contribution.}, Shaolei Liu\footnotemark[1], Manning Wang\thanks{Corresponding Authors}, Zhijian Song\footnotemark[2]}
\affiliations{
    Digital Medical Research Center, School of Basic Medical Science, Fudan University, Shanghai 200032, China\\
    Shanghai Key Lab of Medical Image Computing and Computer Assisted Intervention\\
    \{lhqu20, slliu, mnwang, zjsong\}@fudan.edu.cn
}

\begin{document}

\maketitle

\begin{abstract}
    In this paper, we propose TransMEF, a transformer-based multi-exposure image fusion framework that uses self-supervised multi-task learning. The framework is based on an encoder-decoder network, which can be trained on large natural image datasets and does not require ground truth fusion images. We design three self-supervised reconstruction tasks according to the characteristics of multi-exposure images and conduct these tasks simultaneously using multi-task learning; through this process, the network can learn the characteristics of multi-exposure images and extract more generalized features. In addition, to compensate for the defect in establishing long-range dependencies in CNN-based architectures, we design an encoder that combines a CNN module with a transformer module. This combination enables the network to focus on both local and global information. We evaluated our method and compared it to 11 competitive traditional and deep learning-based methods on the latest released multi-exposure image fusion benchmark dataset, and our method achieved the best performance in both subjective and objective evaluations. Code will be available at \url{https://github.com/miccaiif/TransMEF}.
\end{abstract}

\section{Introduction}
Due to the low dynamic range (LDR) of common imaging sensors, a single image often suffers from underexposure or overexposure and fails to depict the high dynamic range (HDR) of luminance levels in natural scenes. The multi-exposure image fusion (MEF) technique provides an economical and effective solution by fusing LDR images with different exposures into a single HDR image and thus is widely used in HDR imaging for mobile devices \cite{2,3,4}. 

The study of MEF has a long history, and a series of traditional methods have been proposed \cite{5,6,7,8,9,10}. However, their performances are limited because weak hand-crafted representations have low generalizability and are not robust to varying input conditions \cite{1,14,15}.

Recently, deep learning-based algorithms have gradually become mainstream in the MEF field. In these methods, two source images with different exposures are directly input into a fusion network, and the fused image is obtained from the output of the network. The fusion networks can be trained in a common supervised way using ground truth fusion images \cite{14,19,20} or in an unsupervised way by encouraging the fused image to retain different aspects of the important information in the source images \cite{15,16,17,18,21}. However, both supervised and unsupervised MEF methods require a large amount of multi-exposure data for training. Although many researchers \cite{16,64,65} have collected various multi-exposure datasets, their quantities are not comparable to large natural image datasets such as \textit{ImageNet} \cite{22} or \textit{MS-COCO} \cite{52}. The absence of large amounts of training data generally leads to overfitting or tedious parameter optimization. In addition, ground truth is also in high demand for supervised MEF methods but is not commonly available in the field \cite{1}. Some researchers synthesize ground truth images \cite{19} or use the fusion results from other methods as ground truth for training \cite{24,25,26}. However, these ground truth images are not real, and using them leads to inferior performance.

Moreover, all the existing deep learning-based MEF methods utilize convolutional neural networks (CNNs) for feature extraction, but it is difficult for CNNs to model long-range dependencies due to their small receptive field, which is an inherent limitation. In image fusion, the quality of the fused images is related to the pixels within the receptive field as well as to the pixel intensity and texture of the entire image. Therefore, modeling both global and local dependencies is required.

To address the above issues, we propose TransMEF, a transformer-based multi-exposure image fusion framework that uses self-supervised multi-task learning. The framework is based on an encoder-decoder network and is trained on a large natural image dataset using self-supervised image reconstruction to avoid training with multi-exposure images. During the fusion phase, we first apply the trained encoder to extract feature maps from two source images and then apply the trained decoder to generate a fused image from the fused feature maps. We also design three self-supervised reconstruction tasks according to the characteristics of multi-exposure images to train the network so that our network learns these characteristics more effectively.  

In addition, we design an encoder that includes both a CNN module and a transformer module so that the encoder can utilize both local and global information. Extensive experiments demonstrate the effectiveness of the self-supervised reconstruction tasks as well as the transformer-based encoder and show that our method outperforms the state-of-the-art MEF methods in both subjective and objective evaluations.

The main contributions of this paper are summarized as follows:

\begin{itemize}
\item We propose three self-supervised reconstruction tasks according to the characteristics of multi-exposure images and train an encoder-decoder network using multi-task learning so that our network is not only able to be trained on large natural image datasets but also learns the characteristics of multi-exposure images.
\item To compensate for the defect in establishing long-range dependencies in CNN-based architectures, we design an encoder that combines a CNN module with a transformer module, which enables the network to utilize both local and global information during feature extraction.
\item To provide a fair and comprehensive comparison with other fusion methods, we used the latest released multi-exposure image fusion benchmark dataset \cite{1} as the test dataset. We selected 12 objective evaluation metrics from four perspectives and compared our method to 11 competitive traditional and deep learning-based methods in the MEF field. Our method achieves the best performance in both subjective and objective evaluations. 
\end{itemize}

\section{Related Work}

\subsection{Traditional MEF Algorithms}
Traditional MEF methods can be further classified into spatial domain-based fusion methods \cite{6,7,8,9,10} and transform domain-based fusion methods \cite{5,11,12,13}. Spatial domain methods calculate the fused image’s pixel values directly from the source image’s pixel values, and three types of techniques are commonly used to fuse images in the spatial domain, namely, pixel-based methods \cite{6,7}, patch-based methods \cite{8,9} and optimization-based methods \cite{10}. 

In transform domain-based fusion algorithms, the source images are first transformed to a specific transform domain (such as the wavelet domain) to obtain different frequency components, and then appropriate fusion rules are used to fuse different frequency components. Finally, the fused images are obtained by inversely transforming the fused frequency components. Commonly used transform methods include pyramid transform \cite{11}, Laplacian pyramid \cite{12}, wavelet transform \cite{5}, and edge-preserving smoothing \cite{13}, among others.

Although traditional methods achieve promising fusion results, weak hand-crafted representations with low generalizability hinder further improvement.

\subsection{Deep-Learning Based MEF Algorithms}
In deep learning-based algorithms, two source images with different exposures are directly input into a fusion network, and the network outputs the fused image. The fusion networks can be trained with ground truth fusion images \cite{14,19,20} or similarity metric-based loss functions \cite{15,16,17,18,21}.

Due to the lack of real ground truth fusion images in the MEF field, several methods have been proposed to synthesize ground truth. For example, Wang et al. \cite{19} generated ground truth data by changing the pixel intensity of normal images \cite{22}, while other researchers utilized the fusion results from other MEF methods as ground truth \cite{24,25,26}. However, these ground truth images are not real, leading to inferior fusion performance.

In addition to training with ground truth images, another research direction is to train the fusion network with similarity metric-based loss functions that encourage the fusion image to retain important information from different aspects of the source images. For example, Prabhakar et al. \cite{16} applied a no-reference image quality metric (MEF-SSIM) as a loss function. Zhang et al. \cite{18} designed a loss function based on gradient and intensity information to perform unsupervised training. Xu et al. \cite{15,17} proposed U2Fusion, in which a fusion network is trained to preserve the adaptive similarity between the fusion result and the source images. Although these methods do not require ground truth images, a large number of multi-exposure images are still in high demand for training. Although several multi-exposure datasets \cite{16,64,65} have been collected, their quantities are incomparable to large natural image datasets such as \textit{ImageNet} \cite{22} or \textit{MS-COCO} \cite{52}. The absence of large amounts of training data leads to overfitting or tedious parameter optimization.

Notably, researchers have already utilized encoder-decoder networks in infrared and visible image fusion \cite{27} as well as multi-focus image fusion tasks \cite{28}. They trained the encoder-decoder network on the natural image dataset, but the network cannot effectively learn the characteristics of multi-exposure images due to the domain discrepancy. In contrast, we design three self-supervised reconstruction tasks according to the characteristics of multi-exposure images so that our network can not only be trained on large natural image datasets but will also be able to learn the characteristics of multi-exposure images.

\begin{figure*}[htbp]
    \centering
    \includegraphics[scale=0.37]{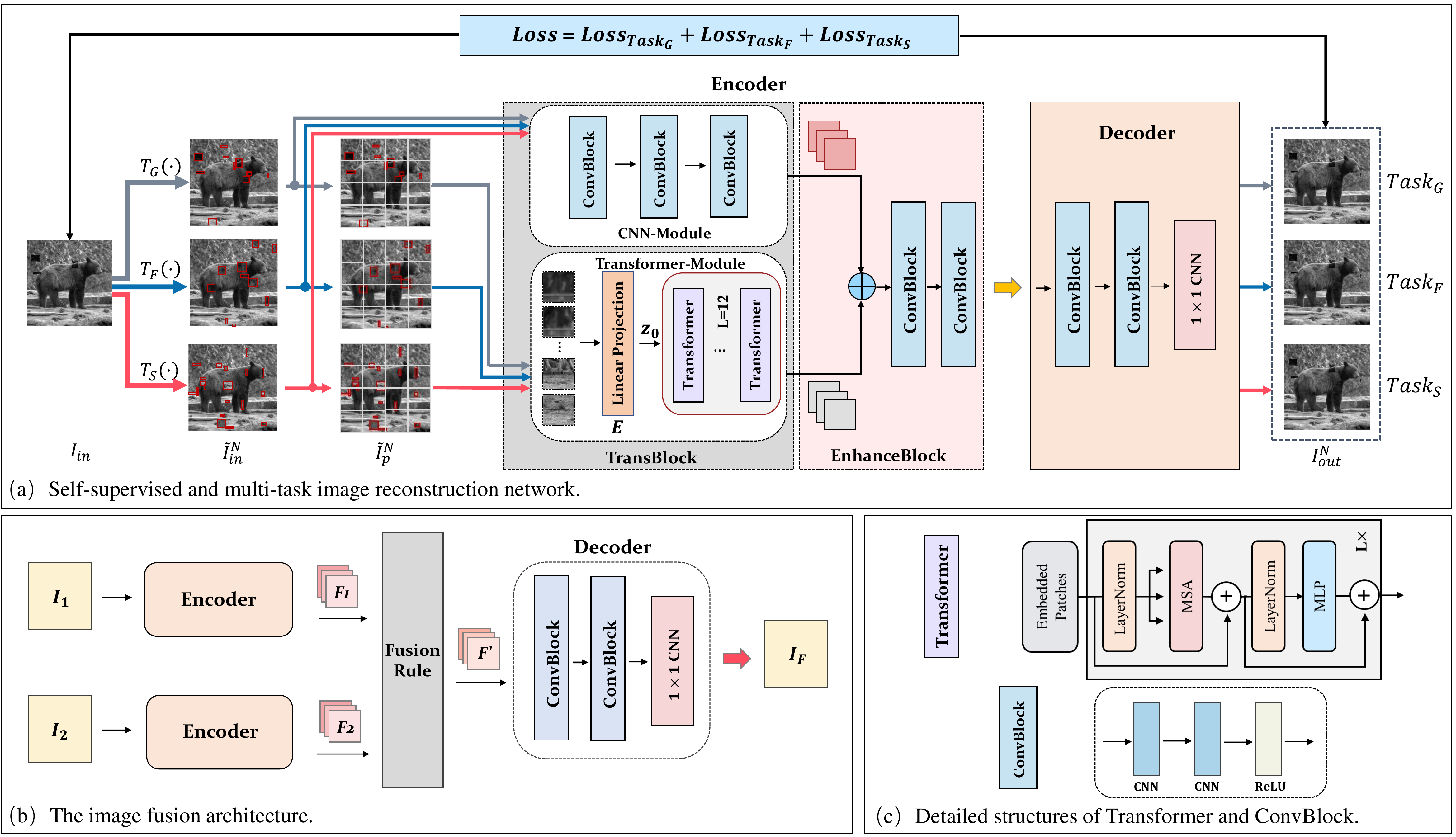}
    \caption{TransMEF image fusion framework. (a) The proposed self-supervised image reconstruction network, which uses multi-task learning. (b) The image fusion architecture. (c) Detailed structures of Transformer and ConvBlock.}
    \label{figure1}
  \end{figure*}

\section{Method}

\subsection{Framework Overview}
As shown in Figure \ref{figure1}, our framework is an encoder-decoder-based architecture. We train the network via image reconstruction on a large natural image dataset. During the fusion phase, we apply the trained encoder to extract feature maps from a pair of source images and then fuse the two feature maps and input it into the decoder to generate the fused image. 

The framework training process is shown in Figure \ref{figure1} (a), where we use the network to perform self-supervised image reconstruction tasks, i.e., to reconstruct the original image from the destroyed input image. Concretely, given an original image ${\ I}_{in\ }\in\mathbb{R}^{H\times W}$, three different destroyed images ${\ \tilde{I}}_{in}^N(N=1,2,3)$ are generated by destroying several subregions using one of the three different transformations (Gamma-based transformation $T_G$(·), Fourier-based transformation $T_F$(·) and global region shuffling $T_S$(·)). The destroyed images are input into the Encoder, which consists of a feature extraction module, TransBlock, and a feature enhancement module, EnhanceBlock. TransBlock uses a CNN module and a transformer module for feature extraction. The destroyed images ${\tilde{I}}_{in}^N$ are directly input into the CNN-Module, and concurrently, they are divided into patches ${\tilde{I}}_p^N$, which are then input into the Transformer-Module. The EnhanceBlock aggregates and enhances the feature maps extracted from the CNN-Module and the Transformer-Module. Finally, the image features extracted by the Encoder are used to obtain the reconstructed image $I_{out}^N(N=1,2,3)$ by the Decoder. We utilize a multi-task learning approach that simultaneously performs three self-supervised reconstruction tasks. The detailed structure of the Encoder and the three reconstruction tasks are introduced in Sections 3.2 and 3.3, respectively.

The trained Encoder and Decoder are then used for image fusion, as shown in Figure \ref{figure1} (b). Specifically, two source images $I_k\ (k=1,2)$ are first input to the Encoder for feature encoding, and then the extracted feature maps $F_1$ and $F_2$ are fused using the Fusion Rule to obtain the fused feature maps $F^\prime$. Finally, the fused image $I_F$ is reconstructed by the Decoder. The fusion rule is described in detail in Section 3.4. Here, we only introduce this framework’s pipeline for single-channel grayscale image fusion, and the fusion of color images is elaborated in Section 3.5.

\subsection{Transformer-Based Encoder-Decoder Framework}

\subsubsection{Encoder-Decoder Framework for Image Reconstruction}

The encoder-decoder network is shown in Figure \ref{figure1} (a). Using a single self-supervised reconstruction task as an example, given a training image ${I}_{in}\in\mathbb{R}^{H \times W}$, we first randomly generate 10 image subregions\ $x_i\in\mathbb{R}^{H_i\times W_i},\left(i=1,2,\ldots10\right)$ to form a set that will be transformed\ $\chi=\{x_1,x_2,\ldots,x_{10}\}$, where the sizes of the subregions $H_i$, $W_i$ are all random values uniformly sampled from the positive integer set [1,25]. After that, we transform each subregion $x_i$ in the set $\chi$ with an image transform tailored for multi-exposure image fusion (the three different transformations are described in detail in Section 3.3) to obtain the set of transformed subregions\ $\widetilde{\chi}$, which are then used to replace the original subregions to obtain the transformed image ${\tilde{I}}_{in}$. In Figure \ref{figure1} (a), $T_G$(·), $T_F$(·) and $T_S$(·) represent the transformation based on Gamma transform, Fourier transform and global region shuffling, respectively.

The Encoder contains a feature extraction module, TransBlock, and a feature enhancement module, EnhanceBlock. The detailed architecture of TransBlock is introduced in the following section. The feature enhancement module, EnhanceBlock, aggregates and enhances the feature maps extracted by TransBlock so that the Encoder can better integrate the global and local features. Concretely, we concatenate the two feature maps from the CNN-Module and the Transformer-Module in TransBlock and input them into two sequentially connected ConvBlock layers to achieve feature enhancement. As shown in Figure \ref{figure1} (c), each ConvBlock consists of two convolutional layers with a kernel size of 3$\times$3, a padding of 1 and one ReLU activation layer. The Decoder contains two sequentially connected ConvBlock layers and a final 1$\times$1 convolution to reconstruct the original image.

\subsubsection{TransBlock: A Powerful Feature Extractor}

Inspired by TransUnet \cite{43} and ViT \cite{30}, we propose a feature extraction module, TransBlock, that combines the CNN and transformer architecture to model both local and global dependencies in images. The architecture of TransBlock is shown in Figure \ref{figure1} (a). Specifically, the CNN-Module consists of three sequentially connected ConvBlock layers, and the input to the CNN-Module is the destroyed images. Simultaneously, the destroyed image ${ \tilde{I}}_{in}\in\mathbb{R}^{H\times W}$\ is divided into a total of M patches of size $\frac{H}{P}\times\frac{W}{P}$. The patches are used to construct the sequence $x_{seq}\in\mathbb{R}^{M\times P^2}$, where $x_{seq}=\left\{x_p^k\right\}$, $\left(k=1,2,\ldots M\right)$,$\ M=HW/P^2$ and  $P$ is the size of the patches. The sequence is fed into the Transformer-Module, which starts with a patch embeddings linear projection E, and the encoded sequence features${\ z}_0\in\mathbb{R}^{M\times D}$ are obtained. Then, ${\ z}_0$ passes through L Transformer layers and the output of each layer is denoted as $z_l(l=1...L)$. Figure \ref{figure1} (c) illustrates the architecture of one Transformer layer, which consists of a multi-head attention mechanism (MSA) block and a multi-layer perceptron (MLP) block, where layer normalization (LN) is applied before every block and residual connections are applied after every block. The MLP block consists of two linear layers with a GELU activation function. 

\subsubsection{Loss Function}
Our architecture applies a multi-task learning approach to simultaneously perform three self-supervised reconstruction tasks using the following loss function.
\begin{equation}
    Loss\ = {Loss}_{{Task}_G}+{Loss}_{{Task}_F}+ {Loss}_{{Task}_S} \label{eq8}
\end{equation}

where $Loss$\ denotes the overall loss function. ${Loss}_{{Task}_G}$, ${Loss}_{{Task}_F}$ and ${Loss}_{{Task}_S}$ are the loss functions of the three self-supervised reconstruction tasks. 

In each reconstruction task, we encourage the network to not only learn the pixel-level image reconstruction, but also capture the structural and gradient information in the image. Therefore, the loss of each reconstruction task contains three parts and is defined as follows:


\begin{equation}
    \mathcal{L} = \mathcal{L}_{mse} + \lambda_1\mathcal{L}_{ssim} + \lambda_2\mathcal{L}_{TV} \label{eq9}
\end{equation}
where $L_{mse}$ is the mean square error (MSE) loss function, $\mathcal{L}_{ssim}$ is the structural similarity (SSIM) loss function, and $\mathcal{L}_{TV}$ is the total variation loss function. $\lambda_1$ and $\lambda_2$ are two hyperparameters that are empirically set to 20.

The MSE loss is used to ensure pixel-level reconstruction and is defined as follows:

\begin{equation}
    \mathcal{L}_{mse}=||I_{out}-I_{in}{||}_2 \label{eq10}
\end{equation}

where $I_{out}$ is the output, a fused image reconstructed by the network, and $I_{in}$ represents the input, the original image.

The SSIM loss helps the model better learn structural information from images and is defined as:
\begin{equation}
    \mathcal{L}_{ssim}=1-SSIM(I_{out},I_{in}) \label{eq11}
\end{equation}
The total variation loss $\mathcal{L}_{TV}$ introduced in VIFNet \cite{44} is used to better preserve gradients in the source images and further eliminate noise. It is defined as follows:
\begin{equation}
    R\left(p,q\right)=I_{out}\left(p,q\right)-I_{in}\left(p,q\right) \label{eq12}
\end{equation}


\begin{align}
    \mathcal{L}_{TV}=\sum_{p,q}{(||R\left(p,q+1\right)-R(p,q){||}_2} \notag\\
    +||R\left(p+1,q\right)-R(p,q){||}_2) \label{eq13}
\end{align}
  
where $R\left(p,q\right)$\ denotes the difference between the original image and the reconstructed image, $||$ · $||_2$ denotes the $l_2$ norm, and $p$, $q$ represent the horizontal and vertical coordinates of the image's pixels, respectively.

\subsection{Three Specific Self-Supervised Image Reconstruction Tasks}
In this section, we introduce three transformations that destroy the original images and generate the input for the image reconstruction encoder-decoder network. An example that shows the image and the corresponding subregions before and after the transformations is presented in Appendices Section A.

\textbf{(1) Learning Scene Content and Luminance Information using Gamma-based Transformation.} In general, overexposed images contain sufficient content and structural information in dark regions, while underexposed images contain sufficient color and structural information in bright regions. In the fused image, it is desirable to maintain uniform brightness while retaining rich information in all regions \cite{15,16}. We adopt Gamma transform to change the luminance in several subregions of the original image and train the network to reconstruct that original image. In this process, our network learns the content and structural information from the images at different luminance levels.

Gamma transform is defined as:
\begin{equation}
    \widetilde{\psi}\ =\ \Gamma\left(\psi\right)=\psi^{gamma} \label{eq14}
\end{equation}
where $\widetilde{\psi}$ and $\psi$ are the transformed and original pixel values, respectively. For each pixel in the selected subregion $x_i$, we use a random Gamma transform $\Gamma$ to change the luminance, where $gamma$ is a random value uniformly sampled from the interval [0, 3]. 

\textbf{(2) Learning Texture and Detail Information using Fourier-based Transformation.} We introduce a self-supervised task based on Fourier transform that enables the network to learn texture and detail information from the frequency domain.

In the discrete Fourier transform (DFT) of an image, the amplitude spectrum determines the image’s intensities, while the phase spectrum primarily determines the high-level semantics of the image and contains information about the image’s content and the location of the objects. (See Appendices Section A.2 for further descriptions and experiments).

Underexposed images are too dark due to insufficient exposure time, and overexposed images are too bright due to a long exposure time, both of which result in inappropriate image intensity distribution. Therefore, it is critical to encourage the network to learn the proper intensity distribution from the images.

Despite the poor intensity distribution in both underexposed and overexposed images, the shape and content of the objects in the image are still well-contained in the phase spectrum. Hence, it is beneficial to build a network that can capture that shape and content information under such circumstances.

To this end, for the selected image subregions, we first perform Fourier transform to obtain the amplitude and phase spectrum. Then, we destroy the subregions in the frequency domain. Specifically, Gaussian blurring $(\sigma=0.5)$ is used to change the amplitude spectrum, and random swapping is performed $n_p$\ times on all phase values in the phase spectrum, where $n_p$ is a random number in the positive integer set [1, 5].

\textbf{(3) Learning Structure and Semantic Information using Global Region Shuffling} We introduce the global region shuffling transform \cite{50} to destroy the original images, thus enabling the network to learn the structure and semantic information through image reconstruction. Specifically, for each image subregion $x_i$\ in\ the set of image subregions $\chi$ selected in the original image ${I}_{in}$, we randomly select another image subregion $x_i^\prime$ with the same size of $x_i$. After that, they are swapped and the process is repeated 10 times to obtain the destroyed image. 

\subsection{Fusion Rule}

Because our network already has a strong feature extraction capability, we simply average the feature maps from the two source images $F_1$ and $F_2$ to obtain the fused feature maps $F^{\prime}$, which is then forwarded to the Decoder. 

\subsection{Managing RGB Input}
We adopt a strategy commonly applied in previous deep learning-based studies to fuse RGB multi-exposure images \cite{18}. The color image’s RGB channels are first converted to the YCbCr color space. Then, the Y (luminance) channel is fused using our network, and the information in the Cb and Cr (chrominance) channels is fused using the traditional weighted average method, defined as:
\begin{equation}
    C_f=\frac{C_1\left(\left|C_1-\tau\right|\right)+C_2(\left|C_2-\tau\right|)}{\left|C_1-\tau\right|+\left|C_2-\tau\right|} \label{eq16}
\end{equation}
where $C_1$ and $C_2$ represent the Cb (or Cr) channel values from the multi-exposure images. $C_f$ denotes their fused channel result, where $\tau$ is set to 128. Finally, the fused Y channel, Cb and Cr channels are converted back to the RGB space.

\begin{figure*}[htbp]
    \centering
    \includegraphics[scale=0.64]{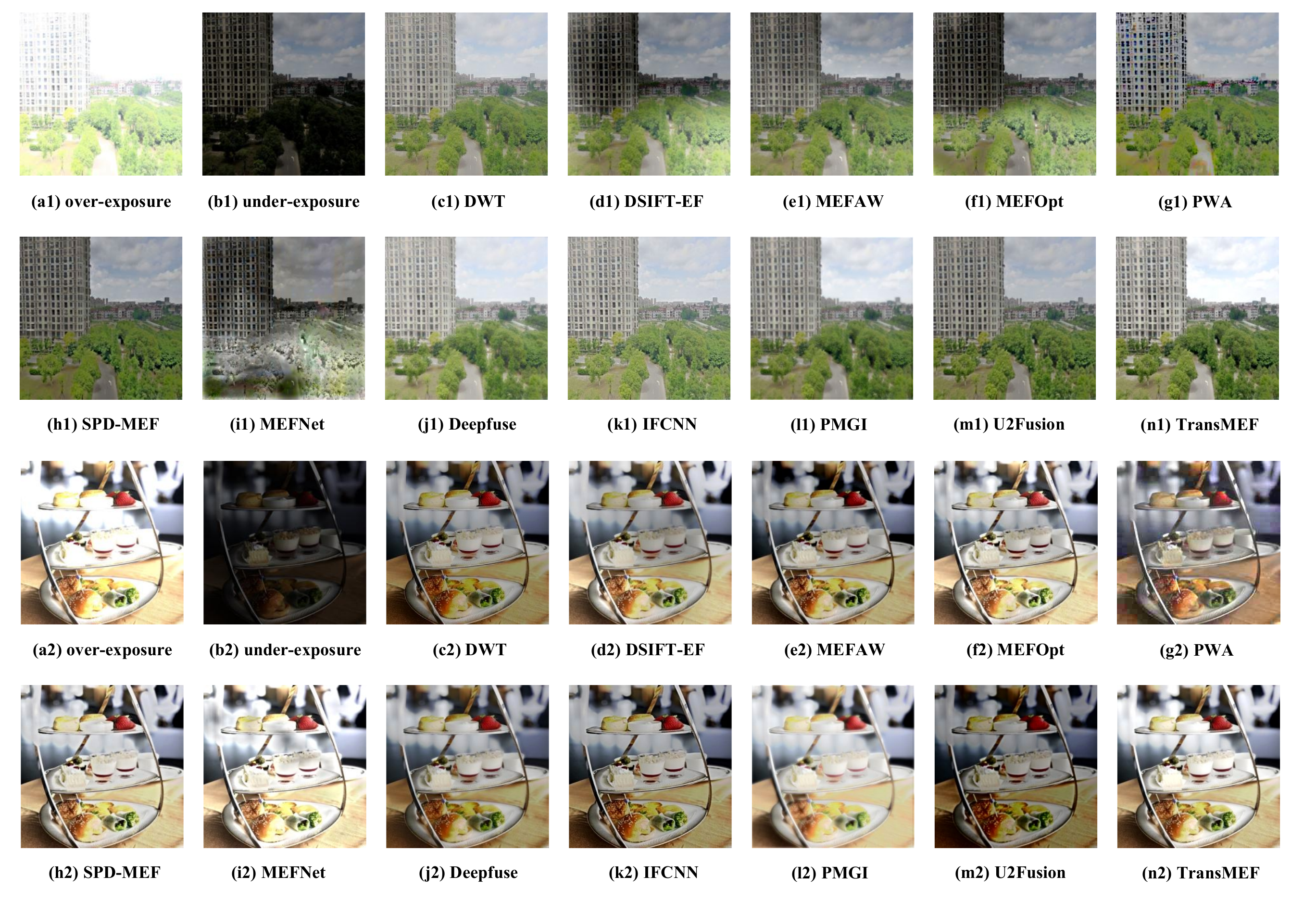}
    \caption{Two examples of source image pairs and fusion results from different methods. (a1)-(b1) and (a2)-(b2) are the source image pairs, and (c1)-(n1) and (c2)-(n2) are the fusion results from various methods.}
    \label{figure2}
  \end{figure*}
  \begin{table*}[htbp]
    \centering
    \caption{Objective evaluation results for the benchmark dataset with the maximum values depicted in red.}
    \includegraphics[scale=0.49]{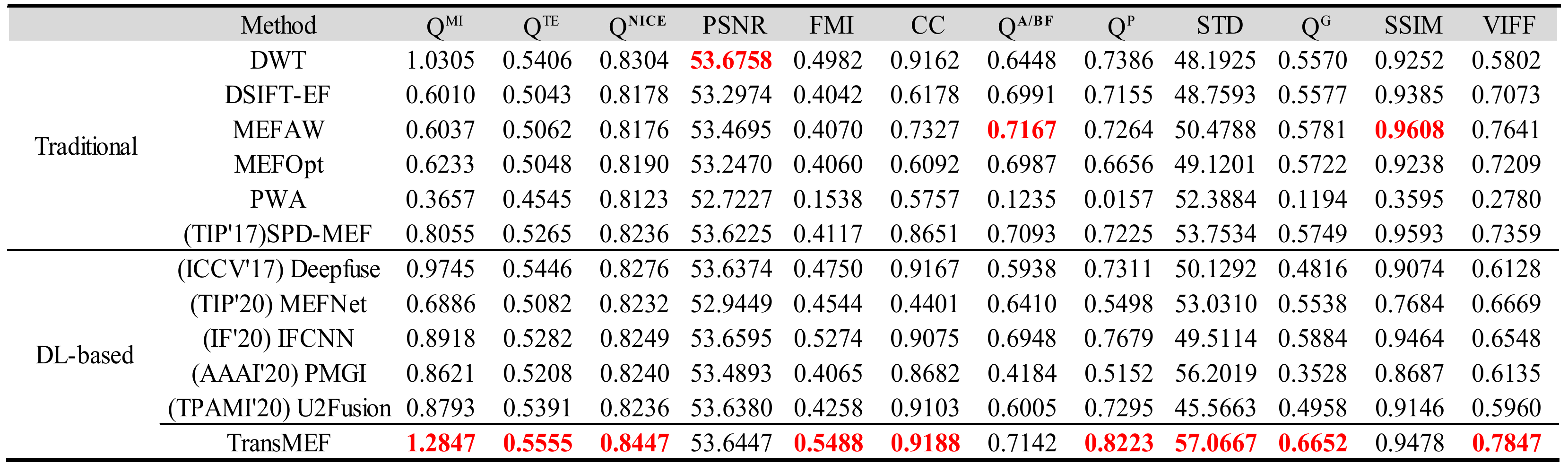}
    \label{table1}
  \end{table*}

\section{Experiments and Results}

\begin{table*}[htbp]
    \centering
    \caption{Results of the ablation study for TransBlock and three self-supervised reconstruction tasks using 20\% of the training data.}
    \includegraphics[scale=0.49]{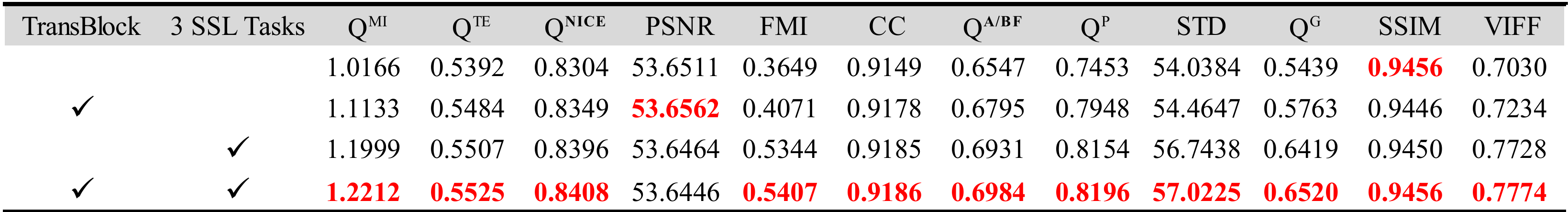}
    \label{table2}
  \end{table*}

\begin{table*}[htbp]
    \centering
    \caption{Results of the ablation study for each self-supervised reconstruction task using 20\% of the training data.}
    \includegraphics[scale=0.49]{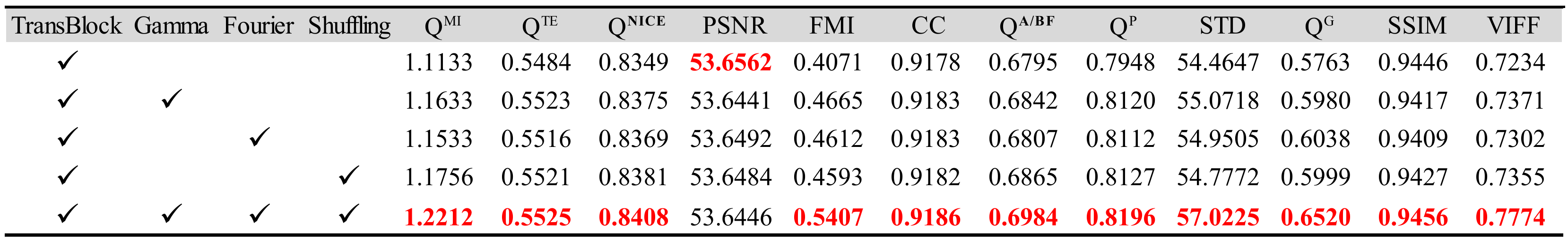}
    \label{table3}
  \end{table*}

\subsection{Datasets}
We used the large natural dataset \textit{MS-COCO} \cite{52} to train the encoder-decoder network. \textit{MS-COCO} contains more than 70,000 natural images of various scenes. For convenience, all images were resized to 256 $\times$ 256 and converted into grayscale images. It is worth mentioning that although many competitive MEF algorithms have been proposed, they are not evaluated on a unified MEF benchmark. We used the latest released multi-exposure image fusion benchmark dataset \cite{1} as the test dataset. This benchmark dataset consists of 100 pairs of multi-exposure images with a variety of scenes and multiple objects.

\subsection{Implementation Details}
Our network was trained on an NVIDIA GTX 3090 GPU with a batch size of 64 and 70 epoch. We used an ADAM optimizer and a cosine annealing learning rate adjustment strategy with a learning rate of 1e-4 and a weight decay of 0.0005. For a 256 $\times$ 256 training image, we randomly generated 10 subregions of random size to form the set $\chi$ to be transformed. In TransBlock, we divided the transformed input image into patches with the size of 16 $\times$ 16 and constructed the sequence $x_{seq}$.

\subsection{Evaluation Metrics}
We rigorously evaluated our method using both subjective and objective evaluations \cite{1}. Subjective evaluation is the observer's subjective assessment of the quality of the fused images in terms of sharpness, detail, and contrast, among other factors. In the objective evaluation, to provide a fair and comprehensive comparison with other fusion methods, we selected 12 objective evaluation metrics from four perspectives. These include information theory-based metrics, $\rm{Q^{MI}}$, $\rm{Q^{TE}}$, $\rm{Q^{NICE}}$, $\rm{PSNR}$, $\rm{FMI}$; image feature-based metrics, $\rm{Q^{A/BF}}$, $\rm{Q^P}$, $\rm{STD}$, $\rm{Q^G}$; image structural similarity-based metrics, $\rm{SSIM}$, $\rm{CC}$; and human perception inspired metrics, $\rm{VIF}$. Details about the metrics can be found in Appendices Section C. All objective metrics are calculated as the average for the 100 fused images, and a larger value indicates better performance for all metrics. 

We compared our method with 11 competitive traditional methods \cite{5,6,7,8,9,10} and deep learning methods \cite{14,15,16,17,18,21} in the MEF field. The comparison methods are as follows: The traditional methods include: DWT \cite{5}, DSIFT-EF \cite{6}, MEFAW \cite{7}, PWA \cite{8}, SPD-MEF \cite{9}, and MEFOpt \cite{10} and the deep learning-based methods include Deepfuse \cite{16}, MEFNet \cite{21}, U2Fusion \cite{15}, PMGI \cite{18}, and IFCNN \cite{14}. 

\subsection{Subjective Evaluation}
Figure \ref{figure2} shows the fusion results from our method and our competitors in an indoor and outdoor scene. More fusion results are shown in Appendices Section D.

When fusing the first pair of source images in Figure \ref{figure2} (a1) and (b1), DSIFT-EF, MEFAW, MEFOpt, SPD-MEF and MEFNet result in disappointing luminance maintenance, and the fused images appear dark. PWA introduces artifacts, and the color is unrealistic. Although DWT, Deepfuse, PMGI, IFCNN and U2Fusion maintain moderate luminance, their fusion results suffer from low contrast and fail to depict the image’s details. In comparison, our method maintains the best luminance and contrast and simultaneously displays excellent details with better visual perception.

When fusing the second pair of source images in Figure \ref{figure2} (a2) and (b2), most methods fail to maintain appropriate luminance. MEFNet and PMGI maintain relatively better luminance but introduce artifacts and blurring. Clearly, our method maintains optimal luminance and contrast and simultaneously retains more detailed information.

\subsection{Objective Evaluation}

Table \ref{table1} presents the objective evaluation for all comparison methods on the benchmark dataset. Our method achieves the best performance for nine of the 12 metrics, while for the other three metrics, the gap between our method’s results and the best results is small.

\section{Ablation Study}

\subsection{Ablation Study for TransBlock}
To verify the effectiveness of TransBlock, we conducted an ablation study using 20\% of the training data, and the results of the ablation study are shown in Table \ref{table2}. Regardless of whether the proposed self-supervised reconstruction tasks are used, adding TransBlock always improves the fusion performance.

To further explain why TransBlock is effective, we visualized the effect of image reconstruction using both the traditional CNN architecture and the model that includes TransBlock. It can be seen that the latter reconstructed better details. More information can be found in Appendices Section B.

\subsection{Ablation Study for Three Specific Self-Supervised Image Reconstruction Tasks}
In this ablation study, we demonstrate the effectiveness of each of the self-supervised reconstruction tasks and the superiority of performing them simultaneously in a multi-task manner. This study was performed using 20\% of the training data, and the experimental results are shown in Table \ref{table3}. The results show that each of the self-supervised reconstruction tasks alone can improve the fusion performance, and the overall best performance is achieved by conducting the three tasks simultaneously through multi-task learning.

\section{Conclusion}

In this paper, we propose TransMEF, a transformer-based multi-exposure image fusion framework via self-supervised multi-task learning. TransMEF is based on an encoder-decoder structure so that it can be trained on large natural image datasets. The TransMEF encoder integrates a CNN module and a transformer module so that the network can focus on both local and global information. In addition, we design three self-supervised reconstruction tasks according to the characteristics of multi-exposure images and conduct these tasks simultaneously using multi-task learning so that the network can learn those characteristics during the process of image reconstruction. Extensive experiments show that our new method achieves state-of-the-art performance when compared with existing competitive methods in both subjective and objective evaluations. The proposed TransBlock and the self-supervised reconstruction tasks have the potential to be applied in other image fusion tasks and other areas of image processing.


\begin{appendices}

\begin{figure*}[htbp]
    \centering
    \includegraphics[scale=0.38]{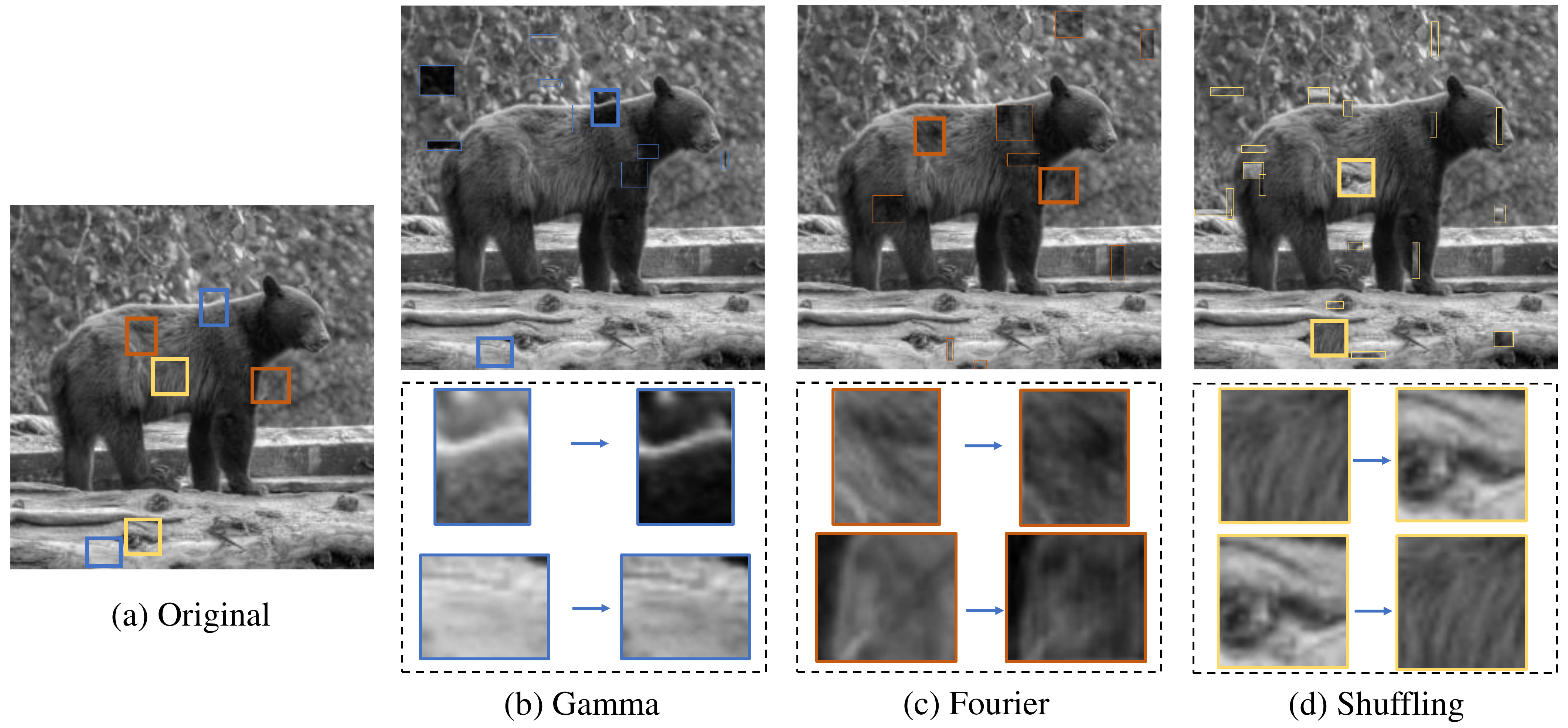}
    \caption{Illustration of three image transformations based on Gamma transform, Fourier transform and global region shuffling designed for multi-exposure image fusion. The boxes of different colors (blue, orange, and yellow) represent the selected subregions for different transformations. For each transformation, zoom of the original and transformed images in two representative subregions are shown in the lower row, and they are marked with bolded box in the original and transformed images. (a) the original image. (b) the transformed image (upper row) and two transformed subregions (lower row) using Gamma-based transformation. (c) the transformed image (upper row) and two transformed subregions (lower row) using Fourier-based transformation. (d) the transformed image (upper row) and two transformed subregions (lower row) using global region shuffling.}
    \label{figure3}
  \end{figure*}

\begin{figure*}[htbp]
    \centering
    \includegraphics[scale=0.37]{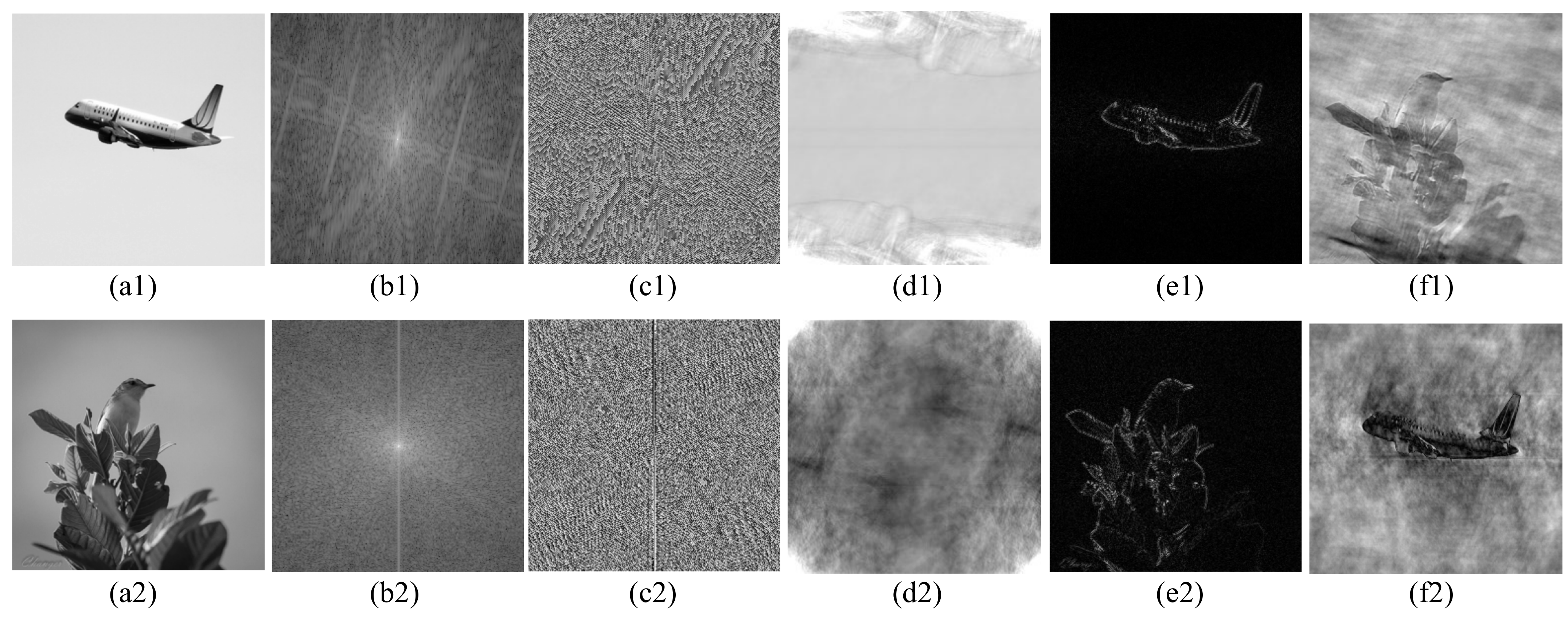}
    \caption{Examples of amplitude-only, phase-only and phase-exchange reconstruction. The original images are shown in (a1-a2) and we performed DFT on them to obtain their amplitude spectrums (b1-b2) and phase spectrums (c1-c2). (d1-d2) are reconstructed images with the corresponding amplitude unchanged but the phase component being set to a constant. (e1-e2) are reconstructed images with the phase unchanged but the amplitude component being set to a constant. (f1-f2) are reconstructed images with the amplitude spectrums unchanged but phase spectrums exchanged.}
    \label{figure4}
  \end{figure*}

\section{Illustration of Three Specific Self-Supervised Image Reconstruction Tasks}
\subsection{Learning Scene Content and Luminance Information using Gamma-based Transformation}
Gamma transform is defined as:
\begin{equation}
    \widetilde{\psi}\ =\ \Gamma\left(\psi\right)=\psi^{gamma} \label{eq1}
\end{equation}


where $\widetilde{\psi}$ and $\psi$ are the transformed and the original pixel values. For each pixel in the selected subregion $x_i$, we perform a random Gamma transformation $\Gamma$ to change the luminance. $gamma$ is a random value uniformly sampled in the interval [0, 3]. Figure \ref{figure3} (b) shows several representative original subregions and corresponding transformed subregions using Gamma-based transformation.

\subsection{Learning Texture and Detail Information using Fourier-based Transformation}
In order to learn the texture and detail information of the source images, a self-supervised task based on 2D discrete Fourier transform (DFT) is introduced in this paper. To illustrate how the phase and amplitude spectrum of DFT influence the appearance of an image, we conducted the following experiments.

Figure \ref{figure4} shows that the amplitude spectrum determines the image's intensities, while the phase spectrum mainly determines the high-level semantics of the image and contains information about the content of the image and the location of the objects \cite{66,67,68,69}. Figure \ref{figure4} (f1-f2) shows that the phase spectrum plays a dominant role in determining the semantics (shape and location of objects) of an image and the amplitude spectrum mainly determines the intensity information. Consequently, we can design Fourier-based self-supervised task to enable the network to learn the texture and detail information from the frequency domain.

For the selected image subregions, we first perform the Fourier transform to obtain the amplitude and the phase spectrum. Then Gaussian blurring ($\sigma$=0.5) is applied to change the amplitude 
spectrum, and randomly swapping is performed $n_p$ times on all phase values in the phase spectrum, where $n_p$ is a random number in the positive integers set [1, 5]. Finally, the network is trained to reconstruct the original image from the destroyed one so that it can learn the texture and detail information from the frequency domain. Figure \ref{figure3} (c) shows several representative original subregions and corresponding transformed subregions using Fourier-based transformation. The specific implementation algorithm is shown in Algorithm \ref{alg1}.
\begin{algorithm}[h]
	\renewcommand{\algorithmicrequire}{\textbf{Input:}}
	\renewcommand{\algorithmicensure}{\textbf{Output:}}
	\caption{Fourier Transformation in a Python-like Style}
	\label{alg1}
	\begin{algorithmic}[1]
        \REQUIRE original image $I_{in}$
        \FOR {$i$ in range(10)}
		    \STATE Randomly select a subregion $x_i\in\mathbb{R}^{H_i\times W_i}$
            \STATE $\rm Spectrum$ = $fft2(x_i)$
            \STATE $\rm Amplitude$ = $abs(\rm Spectrum)$
            \STATE $\rm Phrase$ = $angle(\rm Spectrum)$
            \STATE ${\rm Amplitude}^\prime$ = GaussianBlur($\rm Amplitude$)
            \STATE Generate $n_p$ uniformly distributed over [1,5]
            \FOR {$k$ in range($n_p$)}
            \STATE Randomly shuffling all values in $\rm Phrase$
            \STATE ${\rm Spectrum}^\prime$ = ${\rm Amplitude}^\prime\ast\exp(1j\ast{\rm Phrase}^\prime)$
            \STATE ${\widetilde{x}}_i$=\ ifft2(${\rm Spectrum}^\prime$)
            \ENDFOR
        \ENDFOR
        \ENSURE  transformed image${\ \widetilde{I}}_{in\ }$
	\end{algorithmic}  
\end{algorithm}

\subsection{Learning Structure and Semantic Information using Global Region Shuffling}
\begin{algorithm}[h]
	\renewcommand{\algorithmicrequire}{\textbf{Input:}}
	\renewcommand{\algorithmicensure}{\textbf{Output:}}
	\caption{Global Region Shuffling in a Python-like Style}
	\label{alg2}
	\begin{algorithmic}[1]
        \REQUIRE  original image $I_{in}$
        \FOR {$i$ in range(10)}
		    \STATE Randomly select a subregion $x_i\in\mathbb{R}^{H_i\times W_i}$
            \STATE Randomly select another subregion $x_i^\prime\in\mathbb{R}^{H_i\times W_i}$
            \STATE Swap $x_i$ and $x_i^\prime$
        \ENDFOR
        \ENSURE  transformed image${\ \widetilde{I}}_{in\ }$
	\end{algorithmic}  
\end{algorithm}

Figure \ref{figure3} (d) shows several representative original subregions and corresponding transformed subregions with global region shuffling. The specific implementation algorithm is shown in Algorithm \ref{alg2}.

\section{Comparisons of Image Reconstruction Results with CNN and Transformer}
To address the issue of establishing long-range dependencies in CNN-based architectures, we design an encoder that combines a CNN module with a transformer module, which enables the network to utilize both local and global information during feature extraction. To further explain its effectiveness, we visualized the effect of image reconstruction using both the traditional CNN architecture and the model that includes TransBlock. Two examples of the reconstructed images are shown in Figure \ref{figure5}, and the objective quality metrics of the reconstructed images are shown in Table \ref{table4} below.

\begin{figure}[htbp]
    \centering
    \includegraphics[scale=0.4]{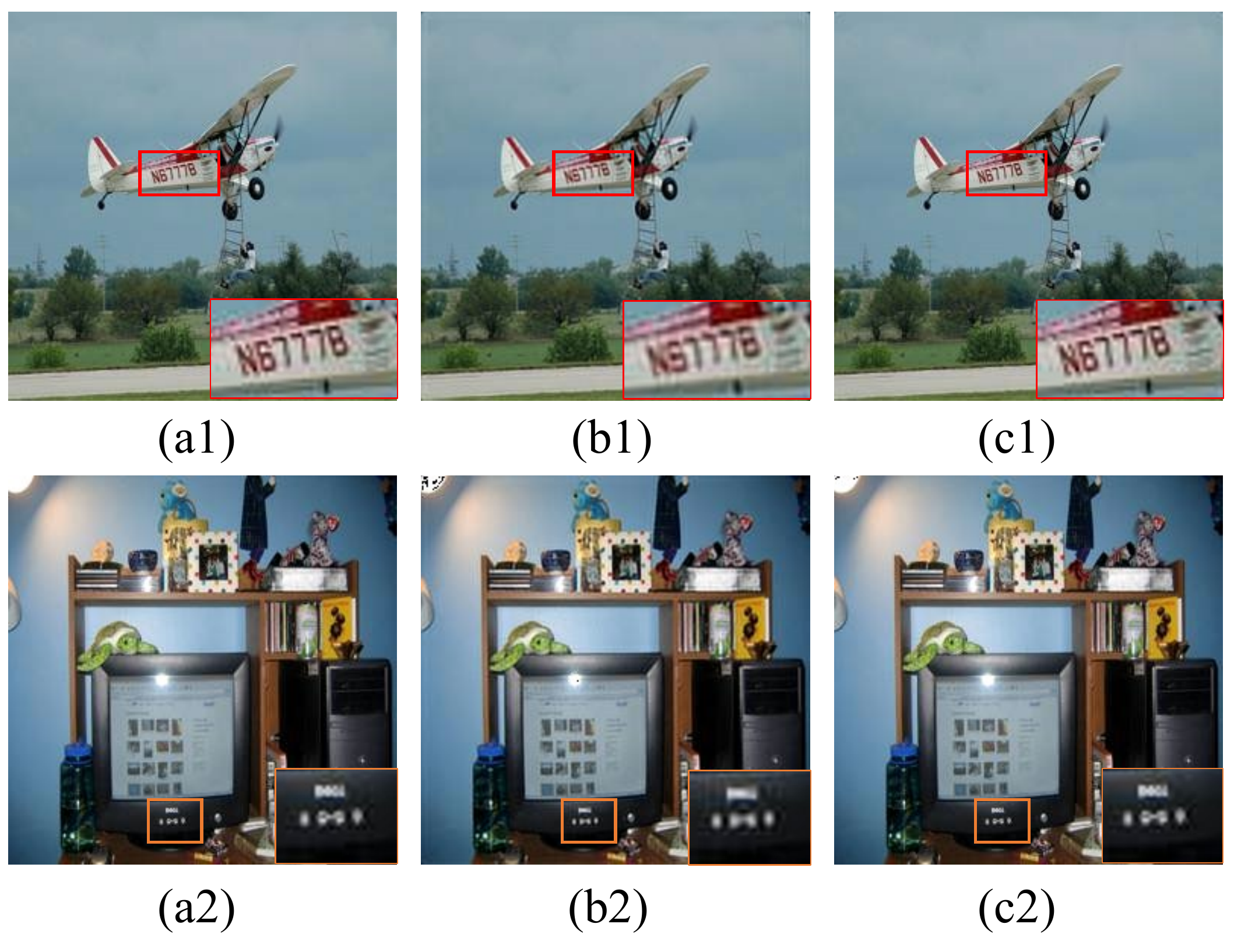}
    \caption{Comparison of the reconstructed images using CNN architecture alone and the model with TransBlock. (a1-a2) show the original images, (b1-b2) show the image reconstruction results with CNN architecture alone, and (c1-c2) show the image reconstruction results of the model using TransBlock.}
    \label{figure5}
  \end{figure}
  
\begin{table}[htbp]
    \centering
    \caption{Quality evaluation of image reconstruction results.}
    \includegraphics[scale=0.57]{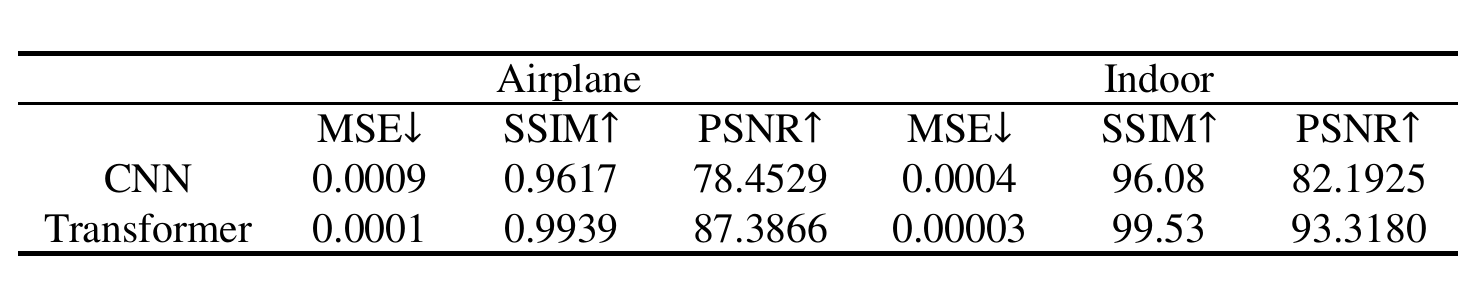}
    \label{table4}
  \end{table}

Figure \ref{figure5} shows that the reconstructed image is more clearly using TransBlock. As the boxed numbers in (a1-c1) and the boxed letters in (a2-c2) show, the CNN reconstruction results are slightly blurred, while the numbers are almost as clear as the original image using TransBlock. In Table \ref{table4}, the mean square error (MSE), structural similarity (SSIM) and peak signal-to-noise ratio (PSNR), which are commonly used as objective evaluation metrics of image quality, are used to evaluate the quality of the reconstructed images. A smaller MSE means more details are preserved from the original images. The larger the SSIM and PSNR metrics, the better the image quality is. Table \ref{table4} illustrates that the MSE of the model using TransBlock is significantly lower than the one with CNN architecture alone. Meanwhile, the SSIM and PSNR metrics are also much higher than the one using CNN architecture alone. These results show that using TransBlock in the Encoder helps improve the quality of the reconstructed images, which indicates the proposed transformer-based Encoder can learn better image features.

\section{Evaluation Metrics}
In the objective evaluation, to provide a fair and comprehensive comparison with other fusion methods, we selected 12 objective evaluation metrics from four perspectives. These include information theory-based metrics, $\rm{Q^{MI}}$ \cite{54}, $\rm{Q^{TE}}$ \cite{55}, $\rm{Q^{NICE}}$ \cite{56}, $\rm{PSNR}$ \cite{57}, $\rm{FMI}$ \cite{58}; image feature-based metrics, $\rm{Q^{A/BF}}$ \cite{59}, $\rm{Q^P}$ \cite{60}, $\rm{STD}$ \cite{61}, $\rm{Q^G}$ \cite{59}; image structural similarity-based metrics, $\rm{SSIM}$ \cite{62}, $\rm{CC}$ \cite{40}; and human perception inspired metrics, $\rm{VIF}$ \cite{63}.

Specifically, $\rm{Q^{MI}}$ measures the amount of information transferred from source images to the fused images. $\rm{Q^{TE}}$ measures the correlation between Tsallis entropy of source images and the fused images. $\rm{Q^{NICE}}$ measures the nonlinear correlation between source images and the fused images. $\rm{PSNR}$ indicates the ratio of peak value power and noise power in the fused images. $\rm{FMI}$ measures the amount of feature information transferred from source images based on mutual information and feature information. $\rm{Q^{A/BF}}$ indicates the amount of edge information transferred from source images to fused images. $\rm{Q^P}$ compares the local cross-correlation of corresponding feature maps of source images and the fused images. $\rm{STD}$ reflects the distribution and contrast of the fused images. $\rm{Q^G}$ measures the amount of edge information transferred from source images to the fused images. $\rm{SSIM}$ reflects the structural similarity between images. $\rm{CC}$ measures the degree of linear correlation of the fused images and the source images. $\rm{VIF}$ measures the information fidelity of the fused images, which is consistent with the human visual system.

\section{More Fusion Results Visualization}
\subsection{Outdoor Scenes}
\begin{figure*}[htbp]
    \centering
    \includegraphics[scale=0.65]{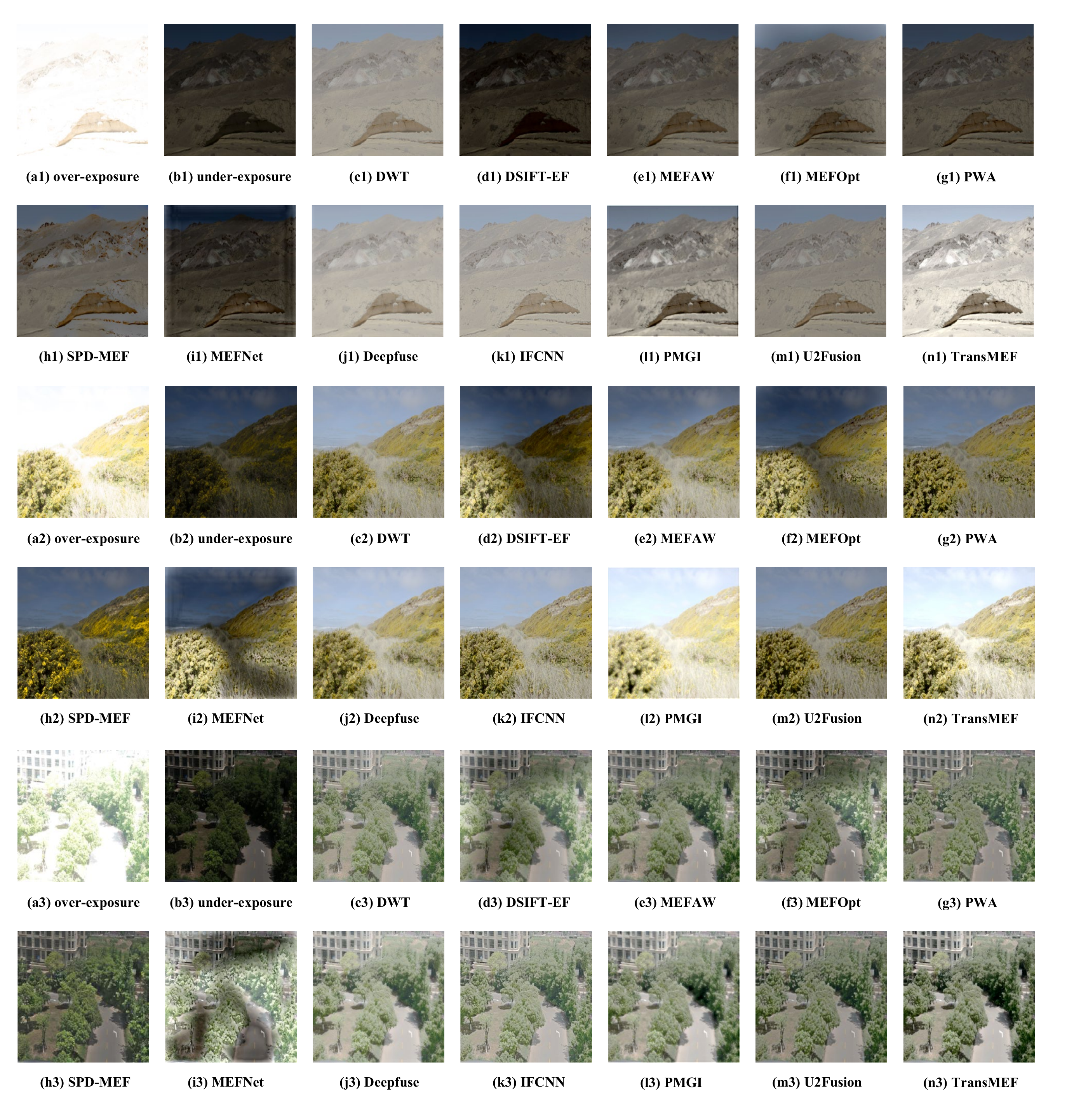}
    \caption{Source images and fusion results of three outdoor scenes of the Benchmark dataset, where (a1)-(b1), (a2)-(b2), (a3)-(b3) are the source image pairs, and the remaining (c1)-(n1), (c2)-(n2), (c3)-(n3) are the fusion results of various methods, respectively.}
    \label{figure6}
  \end{figure*}

Figure \ref{figure6} displays the fusion results from our method and our competitors in three outdoor scenes. When fusing the first pair of source images in Figure \ref{figure6} (a1) and (b1), DSIFT-EF, MEFAW, MEFOpt, PWA, SPD-MEF, MEFNet and U2Fusion result in disappointing luminance maintenance, and the fused images are dark. Although DWT, Deepfuse, and IFCNN maintain moderate luminance, the images suffer from low contrast and fail to show the image's details clearly. PMGI provides better contrast, while its color and luminance are dim. In comparison, our method retains the optimal luminance and contrast, and simultaneously displays excellent details with better visual perception.

When fusing the second pair of source images in Figure \ref{figure6} (a2) and (b2), most of the methods fail to maintain good luminance, and the fused images show artifacts. PMGI retains better luminance, but it shows unclear details.  In comparison, our method preserves the best luminance and contrast, and retains more detailed information.

When fusing the last pair of source images in Figure \ref{figure6} (a3) and (b3), DSIFT-EF, MEFAW, MEFOpt, PWA, SPD-MEF, and MEFNet show severe artifacts in the fused images. Although DWT, Deepfuse, IFCNN, PMGI, and U2Fusion maintain relatively moderate luminance, the contrast of the fused images is low, and detail information is lost. In comparison, our method maintains the optimal luminance and contrast, and simultaneously presents sufficient details.

\subsection{Indoor Scenes}
\begin{figure*}[htbp]
    \centering
    \includegraphics[scale=0.55]{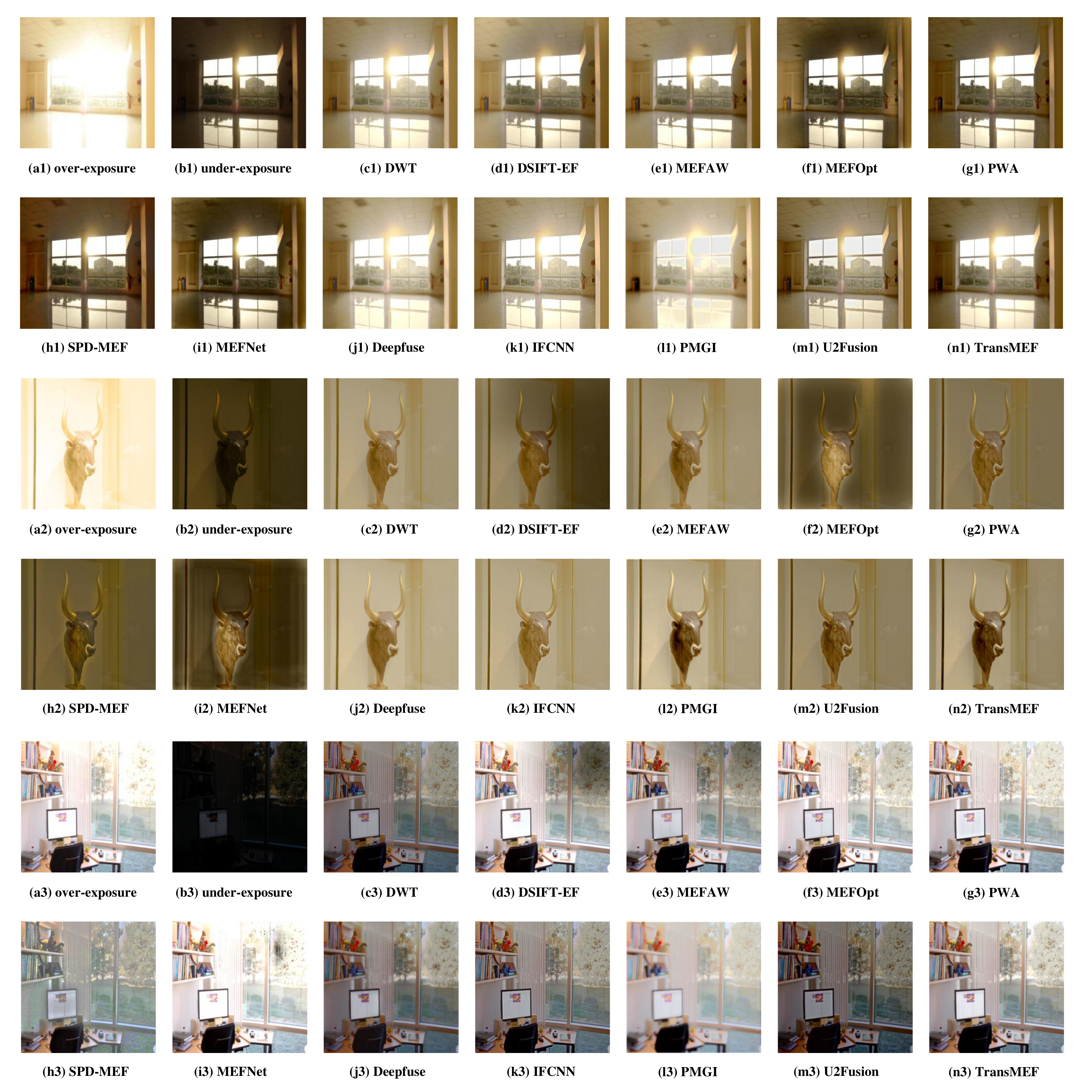}
    \caption{Source images and fusion results of three indoor scenes of the Benchmark dataset, where (a1)-(b1), (a2)-(b2), (a3)-(b3) are the source image pairs, and the remaining (c1)-(n1), (c2)-(n2), (c3)-(n3) are the fusion results of various methods, respectively.}
    \label{figure7}
  \end{figure*}

Figure \ref{figure7} shows the fusion results from our method and our competitors in three indoor scenes. When fusing the first pair of source images in Figure \ref{figure7} (a1) and (b1), MEFOpt, PWA, SPD-MEF, and MEFNet fail to maintain good luminance, and the fused images appear dim. Although the fused images of DWT, Deepfuse, and PMGI are lighter, they remain overexposed in the sunlight. In comparison, DSIFT-EF, MEFAW and IFCNN are able to retain relatively moderate luminance, but the contrast of the images is poor and details are lost. U2Fusion and our method achieve the relatively best visual perception.

When fusing the second pair of source images in Figure \ref{figure7} (a2) and (b2), DSIFT-EF, MEFOpt, SPD-MEF, and MEFNet introduce severe artifacts. DWT, MEFAW, PWA, and U2Fusion are not capable to perform good luminance maintenance, and the fused images are dim. Although Deepfuse and IFCNN maintain moderate luminance, their images have low contrast and lose detail information. In contrast, PMGI and our method maintain the optimal luminance and contrast, while preserving more details.

When fusing the last pair of source images in Figure \ref{figure7} (a3) and (b3) , DSIFT-EF, MEFAW and MEFNet introduce severe artifacts. DWT, SPD-MEF, Deepfuse, IFCNN and U2Fusion cannot perform good luminance retention and the fused images appear dim. MEFOpt, PWA, and PMGI maintain the proper luminance, but their images have low contrast and lose detail information. In comparison, our method retains the optimal luminance and contrast with more details.

\end{appendices}

\section*{Acknowledgments}

This work was supported by National Natural Science Foundation of China under Grant 82072021.

\bibliography{aaai22}

\begin{thebibliography}{49}
\providecommand{\natexlab}[1]{#1}

\bibitem[{Burt and Kolczynski(1993)}]{11}
Burt, P.~J.; and Kolczynski, R.~J. 1993.
\newblock Enhanced image capture through fusion.
\newblock In \emph{1993 (4th) International Conference on Computer Vision
  (ICCV)}, 173--182. IEEE.

\bibitem[{Cai, Gu, and Zhang(2018)}]{64}
Cai, J.; Gu, S.; and Zhang, L. 2018.
\newblock Learning a deep single image contrast enhancer from multi-exposure
  images.
\newblock \emph{IEEE Transactions on Image Processing}, 27(4): 2049--2062.

\bibitem[{Chen et~al.(2021)Chen, Lu, Yu, Luo, Adeli, Wang, Lu, Yuille, and
  Zhou}]{43}
Chen, J.; Lu, Y.; Yu, Q.; Luo, X.; Adeli, E.; Wang, Y.; Lu, L.; Yuille, A.~L.;
  and Zhou, Y. 2021.
\newblock Transunet: Transformers make strong encoders for medical image
  segmentation.
\newblock \emph{arXiv preprint arXiv:2102.04306}.

\bibitem[{Chen and Chuang(2020)}]{25}
Chen, S.-Y.; and Chuang, Y.-Y. 2020.
\newblock Deep exposure fusion with deghosting via homography estimation and
  attention learning.
\newblock In \emph{ICASSP 2020-2020 IEEE International Conference on Acoustics,
  Speech and Signal Processing (ICASSP)}, 1464--1468. IEEE.

\bibitem[{Cvejic, Canagarajah, and Bull(2006)}]{55}
Cvejic, N.; Canagarajah, C.; and Bull, D. 2006.
\newblock Image fusion metric based on mutual information and Tsallis entropy.
\newblock \emph{Electronics letters}, 42(11): 626--627.

\bibitem[{Deng et~al.(2009)Deng, Dong, Socher, Li, Li, and Fei-Fei}]{22}
Deng, J.; Dong, W.; Socher, R.; Li, L.-J.; Li, K.; and Fei-Fei, L. 2009.
\newblock Imagenet: A large-scale hierarchical image database.
\newblock In \emph{2009 IEEE Conference on Computer Vision and Pattern
  Recognition (CVPR)}, 248--255. IEEE.

\bibitem[{Dosovitskiy et~al.(2020)Dosovitskiy, Beyer, Kolesnikov, Weissenborn,
  Zhai, Unterthiner, Dehghani, Minderer, Heigold, Gelly et~al.}]{30}
Dosovitskiy, A.; Beyer, L.; Kolesnikov, A.; Weissenborn, D.; Zhai, X.;
  Unterthiner, T.; Dehghani, M.; Minderer, M.; Heigold, G.; Gelly, S.; et~al.
  2020.
\newblock An Image is Worth 16x16 Words: Transformers for Image Recognition at
  Scale.
\newblock In \emph{International Conference on Learning Representations
  (ICLR)}.

\bibitem[{Haghighat and Razian(2014)}]{58}
Haghighat, M.; and Razian, M.~A. 2014.
\newblock Fast-FMI: Non-reference image fusion metric.
\newblock In \emph{2014 IEEE 8th International Conference on Application of
  Information and Communication Technologies (AICT)}, 1--3. IEEE.

\bibitem[{Han et~al.(2013)Han, Cai, Cao, and Xu}]{63}
Han, Y.; Cai, Y.; Cao, Y.; and Xu, X. 2013.
\newblock A new image fusion performance metric based on visual information
  fidelity.
\newblock \emph{Information fusion}, 14(2): 127--135.

\bibitem[{Hasinoff et~al.(2016)Hasinoff, Sharlet, Geiss, Adams, Barron, Kainz,
  Chen, and Levoy}]{3}
Hasinoff, S.~W.; Sharlet, D.; Geiss, R.; Adams, A.; Barron, J.~T.; Kainz, F.;
  Chen, J.; and Levoy, M. 2016.
\newblock Burst photography for high dynamic range and low-light imaging on
  mobile cameras.
\newblock \emph{ACM Transactions on Graphics}, 35(6): 1--12.

\bibitem[{Hossny, Nahavandi, and Creighton(2008)}]{54}
Hossny, M.; Nahavandi, S.; and Creighton, D. 2008.
\newblock Comments on'Information measure for performance of image fusion'.
\newblock \emph{Electronics letters}, 44(18): 1066--1067.

\bibitem[{Hou et~al.(2020)Hou, Zhou, Nie, Liu, Xiong, Guo, and Yu}]{44}
Hou, R.; Zhou, D.; Nie, R.; Liu, D.; Xiong, L.; Guo, Y.; and Yu, C. 2020.
\newblock VIF-Net: an unsupervised framework for infrared and visible image
  fusion.
\newblock \emph{IEEE Transactions on Computational Imaging}, 6: 640--651.

\bibitem[{Jagalingam and Hegde(2015)}]{57}
Jagalingam, P.; and Hegde, A.~V. 2015.
\newblock A review of quality metrics for fused image.
\newblock \emph{Aquatic Procedia}, 4: 133--142.

\bibitem[{Kang et~al.(2017)Kang, Dong, Zheng, and Yang}]{50}
Kang, G.; Dong, X.; Zheng, L.; and Yang, Y. 2017.
\newblock Patchshuffle regularization.
\newblock \emph{arXiv preprint arXiv:1707.07103}.

\bibitem[{Kou et~al.(2017)Kou, Li, Wen, and Chen}]{13}
Kou, F.; Li, Z.; Wen, C.; and Chen, W. 2017.
\newblock Multi-scale exposure fusion via gradient domain guided image
  filtering.
\newblock In \emph{2017 IEEE International Conference on Multimedia and Expo
  (ICME)}, 1105--1110. IEEE.

\bibitem[{Lee, Park, and Cho(2018)}]{7}
Lee, S.-h.; Park, J.~S.; and Cho, N.~I. 2018.
\newblock A multi-exposure image fusion based on the adaptive weights
  reflecting the relative pixel intensity and global gradient.
\newblock In \emph{2018 25th IEEE International Conference on Image Processing
  (ICIP)}, 1737--1741. IEEE.

\bibitem[{Li, Manjunath, and Mitra(1995)}]{5}
Li, H.; Manjunath, B.; and Mitra, S.~K. 1995.
\newblock Multisensor image fusion using the wavelet transform.
\newblock \emph{Graphical Models and Image Processing}, 57(3): 235--245.

\bibitem[{Li and Wu(2018)}]{27}
Li, H.; and Wu, X.-J. 2018.
\newblock DenseFuse: A fusion approach to infrared and visible images.
\newblock \emph{IEEE Transactions on Image Processing}, 28(5): 2614--2623.

\bibitem[{Li and Zhang(2018)}]{20}
Li, H.; and Zhang, L. 2018.
\newblock Multi-exposure fusion with CNN features.
\newblock In \emph{2018 25th IEEE International Conference on Image Processing
  (ICIP)}, 1723--1727. IEEE.

\bibitem[{Lin et~al.(2014)Lin, Maire, Belongie, Hays, Perona, Ramanan,
  Doll{\'a}r, and Zitnick}]{52}
Lin, T.-Y.; Maire, M.; Belongie, S.; Hays, J.; Perona, P.; Ramanan, D.;
  Doll{\'a}r, P.; and Zitnick, C.~L. 2014.
\newblock Microsoft coco: Common objects in context.
\newblock In \emph{European Conference on Computer Vision (ECCV)}, 740--755.
  Springer.

\bibitem[{Liu and Wang(2015)}]{6}
Liu, Y.; and Wang, Z. 2015.
\newblock Dense SIFT for ghost-free multi-exposure fusion.
\newblock \emph{Journal of Visual Communication and Image Representation}, 31:
  208--224.

\bibitem[{Ma et~al.(2021)Ma, Zhu, Yin, Ban, Huang, and Mukeshimana}]{28}
Ma, B.; Zhu, Y.; Yin, X.; Ban, X.; Huang, H.; and Mukeshimana, M. 2021.
\newblock SESF-fuse: An unsupervised deep model for multi-focus image fusion.
\newblock \emph{Neural Computing and Applications}, 33(11): 5793--5804.

\bibitem[{Ma et~al.(2019{\natexlab{a}})Ma, Yu, Liang, Li, and Jiang}]{40}
Ma, J.; Yu, W.; Liang, P.; Li, C.; and Jiang, J. 2019{\natexlab{a}}.
\newblock FusionGAN: A generative adversarial network for infrared and visible
  image fusion.
\newblock \emph{Information Fusion}, 48: 11--26.

\bibitem[{Ma et~al.(2017{\natexlab{a}})Ma, Duanmu, Yeganeh, and Wang}]{10}
Ma, K.; Duanmu, Z.; Yeganeh, H.; and Wang, Z. 2017{\natexlab{a}}.
\newblock Multi-exposure image fusion by optimizing a structural similarity
  index.
\newblock \emph{IEEE Transactions on Computational Imaging}, 4(1): 60--72.

\bibitem[{Ma et~al.(2019{\natexlab{b}})Ma, Duanmu, Zhu, Fang, and Wang}]{21}
Ma, K.; Duanmu, Z.; Zhu, H.; Fang, Y.; and Wang, Z. 2019{\natexlab{b}}.
\newblock Deep guided learning for fast multi-exposure image fusion.
\newblock \emph{IEEE Transactions on Image Processing}, 29: 2808--2819.

\bibitem[{Ma et~al.(2017{\natexlab{b}})Ma, Li, Yong, Wang, Meng, and Zhang}]{9}
Ma, K.; Li, H.; Yong, H.; Wang, Z.; Meng, D.; and Zhang, L. 2017{\natexlab{b}}.
\newblock Robust multi-exposure image fusion: a structural patch decomposition
  approach.
\newblock \emph{IEEE Transactions on Image Processing}, 26(5): 2519--2532.

\bibitem[{Ma and Wang(2015)}]{8}
Ma, K.; and Wang, Z. 2015.
\newblock Multi-exposure image fusion: A patch-wise approach.
\newblock In \emph{2015 IEEE International Conference on Image Processing
  (ICIP)}, 1717--1721. IEEE.

\bibitem[{Ma, Zeng, and Wang(2015)}]{62}
Ma, K.; Zeng, K.; and Wang, Z. 2015.
\newblock Perceptual quality assessment for multi-exposure image fusion.
\newblock \emph{IEEE Transactions on Image Processing}, 24(11): 3345--3356.

\bibitem[{Mertens, Kautz, and Van~Reeth(2007)}]{12}
Mertens, T.; Kautz, J.; and Van~Reeth, F. 2007.
\newblock Exposure fusion.
\newblock In \emph{15th Pacific Conference on Computer Graphics and
  Applications (PCCGA)}, 382--390. IEEE.

\bibitem[{Oppenheim and Lim(1981)}]{66}
Oppenheim, A.~V.; and Lim, J.~S. 1981.
\newblock The importance of phase in signals.
\newblock \emph{Proceedings of the IEEE}, 69(5): 529--541.

\bibitem[{Piotrowski and Campbell(1982)}]{67}
Piotrowski, L.~N.; and Campbell, F.~W. 1982.
\newblock A demonstration of the visual importance and flexibility of
  spatial-frequency amplitude and phase.
\newblock \emph{Perception}, 11(3): 337--346.

\bibitem[{Ram~Prabhakar, Sai~Srikar, and Venkatesh~Babu(2017)}]{16}
Ram~Prabhakar, K.; Sai~Srikar, V.; and Venkatesh~Babu, R. 2017.
\newblock Deepfuse: A deep unsupervised approach for exposure fusion with
  extreme exposure image pairs.
\newblock In \emph{IEEE International Conference on Computer Vision (ICCV)},
  4714--4722.

\bibitem[{Rao(1997)}]{61}
Rao, Y.-J. 1997.
\newblock In-fibre Bragg grating sensors.
\newblock \emph{Measurement science and technology}, 8(4): 355.

\bibitem[{Reinhard et~al.(2010)Reinhard, Heidrich, Debevec, Pattanaik, Ward,
  and Myszkowski}]{2}
Reinhard, E.; Heidrich, W.; Debevec, P.; Pattanaik, S.; Ward, G.; and
  Myszkowski, K. 2010.
\newblock \emph{High dynamic range imaging: acquisition, display, and
  image-based lighting}.
\newblock Morgan Kaufmann.

\bibitem[{Shen et~al.(2011)Shen, Cheng, Shi, and Basu}]{4}
Shen, R.; Cheng, I.; Shi, J.; and Basu, A. 2011.
\newblock Generalized random walks for fusion of multi-exposure images.
\newblock \emph{IEEE Transactions on Image Processing}, 20(12): 3634--3646.

\bibitem[{Wang et~al.(2018)Wang, Wang, Xu, and Liu}]{19}
Wang, J.; Wang, W.; Xu, G.; and Liu, H. 2018.
\newblock End-to-end exposure fusion using convolutional neural network.
\newblock \emph{IEICE Transactions on Information and Systems}, 101(2):
  560--563.

\bibitem[{Wang, Shen, and Jin(2008)}]{56}
Wang, Q.; Shen, Y.; and Jin, J. 2008.
\newblock Performance evaluation of image fusion techniques.
\newblock \emph{Image fusion: algorithms and applications}, 19: 469--492.

\bibitem[{Xu et~al.(2020{\natexlab{a}})Xu, Ma, Jiang, Guo, and Ling}]{15}
Xu, H.; Ma, J.; Jiang, J.; Guo, X.; and Ling, H. 2020{\natexlab{a}}.
\newblock U2Fusion: A Unified Unsupervised Image Fusion Network.
\newblock \emph{IEEE Transactions on Pattern Analysis and Machine
  Intelligence}.

\bibitem[{Xu et~al.(2020{\natexlab{b}})Xu, Ma, Le, Jiang, and Guo}]{17}
Xu, H.; Ma, J.; Le, Z.; Jiang, J.; and Guo, X. 2020{\natexlab{b}}.
\newblock Fusiondn: A unified densely connected network for image fusion.
\newblock In \emph{Proceedings of the AAAI Conference on Artificial
  Intelligence}, volume~34, 12484--12491.

\bibitem[{Xu, Ma, and Zhang(2020)}]{26}
Xu, H.; Ma, J.; and Zhang, X.-P. 2020.
\newblock MEF-GAN: Multi-exposure image fusion via generative adversarial
  networks.
\newblock \emph{IEEE Transactions on Image Processing}, 29: 7203--7216.

\bibitem[{Xu et~al.(2021)Xu, Zhang, Zhang, Wang, and Tian}]{68}
Xu, Q.; Zhang, R.; Zhang, Y.; Wang, Y.; and Tian, Q. 2021.
\newblock A Fourier-based Framework for Domain Generalization.
\newblock In \emph{Proceedings of the IEEE/CVF Conference on Computer Vision
  and Pattern Recognition}, 14383--14392.

\bibitem[{Xydeas, , and Petrovic(2000)}]{59}
Xydeas, C.; ; and Petrovic, V. 2000.
\newblock Objective image fusion performance measure.
\newblock \emph{Electronics letters}, 36(4): 308--309.

\bibitem[{Yang and Soatto(2020)}]{69}
Yang, Y.; and Soatto, S. 2020.
\newblock Fda: Fourier domain adaptation for semantic segmentation.
\newblock In \emph{Proceedings of the IEEE/CVF Conference on Computer Vision
  and Pattern Recognition}, 4085--4095.

\bibitem[{Yin et~al.(2020)Yin, Chen, Peng, and Tsai}]{24}
Yin, J.-L.; Chen, B.-H.; Peng, Y.-T.; and Tsai, C.-C. 2020.
\newblock Deep prior guided network for high-quality image fusion.
\newblock In \emph{2020 IEEE International Conference on Multimedia and Expo
  (ICME)}, 1--6. IEEE.

\bibitem[{Zeng et~al.(2014)Zeng, Ma, Hassen, and Wang}]{65}
Zeng, K.; Ma, K.; Hassen, R.; and Wang, Z. 2014.
\newblock Perceptual evaluation of multi-exposure image fusion algorithms.
\newblock In \emph{2014 Sixth International Workshop on Quality of Multimedia
  Experience (QoMEX)}, 7--12. IEEE.

\bibitem[{Zhang et~al.(2020{\natexlab{a}})Zhang, Xu, Xiao, Guo, and Ma}]{18}
Zhang, H.; Xu, H.; Xiao, Y.; Guo, X.; and Ma, J. 2020{\natexlab{a}}.
\newblock Rethinking the image fusion: A fast unified image fusion network
  based on proportional maintenance of gradient and intensity.
\newblock In \emph{Proceedings of the AAAI Conference on Artificial
  Intelligence}, volume~34, 12797--12804.

\bibitem[{Zhang(2021)}]{1}
Zhang, X. 2021.
\newblock Benchmarking and comparing multi-exposure image fusion algorithms.
\newblock \emph{Information Fusion}, 74: 111--131.

\bibitem[{Zhang et~al.(2020{\natexlab{b}})Zhang, Liu, Sun, Yan, Zhao, and
  Zhang}]{14}
Zhang, Y.; Liu, Y.; Sun, P.; Yan, H.; Zhao, X.; and Zhang, L.
  2020{\natexlab{b}}.
\newblock IFCNN: A general image fusion framework based on convolutional neural
  network.
\newblock \emph{Information Fusion}, 54: 99--118.

\bibitem[{Zhao, Laganiere, and Liu(2007)}]{60}
Zhao, J.; Laganiere, R.; and Liu, Z. 2007.
\newblock Performance assessment of combinative pixel-level image fusion based
  on an absolute feature measurement.
\newblock \emph{International Journal of Innovative Computing, Information and
  Control}, 3(6): 1433--1447.

\end{thebibliography}
\end{document}